# Energy Decay Network (EDeN)

**Jamie Nicholas Shelley**

Optishell consultancy

## ABSTRACT

This paper and accompanying Python/C++ Framework is the product of the authors perceived problems with narrow (Discrimination based) AI. (Artificial Intelligence) The Framework attempts to develop a genetic transfer of experience through potential structural expressions using a common regulation/exchange value ('energy') to create a model whereby neural architecture and all unit processes are co-dependently developed by genetic and real time signal processing influences; successful routes are defined by stability of the spike distribution per epoch which is influenced by genetically encoded morphological development biases.

These principles are aimed towards creating a diverse and robust network that is capable of adapting to general tasks by training within a simulation designed for transfer learning to other mediums at scale.

Contents

## INTRODUCTION

Sections '*Genetics and genetic algorithms*' '*Nature of information and complexity*' and 'Artificial and biological neurons' are the authors observations and comparisons of neural computing from varying perspectives, this attempts to explain the reasoning behind the EDeN framework development. ' *EDeN Framework and core process overview*' details application of this conjecture to a reduced cycle of operations designed to create a network of 'behavior driven intelligence'.

The section 'Artificial and Biological Neurons' details a neuron model ('Process node') that is evaluated by a common exchange value 'Stability index' which is assigned as a result of how well the node can manage energy locally over training ( influenced by product of historically successful morphology changes that are genetically encoded ('Functome') ).

## INTUITION

I. The assumption that a neuron (process node) competes to survive in return of 'being a good signal processor' by which information can be dimensionally reduced and modeled. Mathematically this is the attempt to remove dependency on a global minimization function , replacing it with behavior that is translated to each unit differently depending on location and required processes of it's own 'survival', separate from global entity training objectives.

II. The morphology and signal processing properties of the network are created from common principles/rules (as opposed to CNN architectures where architecture is manually defined in specialized layers) **[Ref 11]**.

III. Genetics ('Functome') is expressed as morphological biases from internal environmental evaluations; ensuring a relationship between all development steps. This provides a mechanism for internally reasoned structural and functional definitions that are recorded for further cross domain utilisation and intergenerational expression.

IV. As dependency of a high entropy structure increases. As does a need for energy efficiency of it's operation. Once critical boundaries of this operation is met, structural representation of this process is maximized.

### *A high level diagram of the developmental process:*

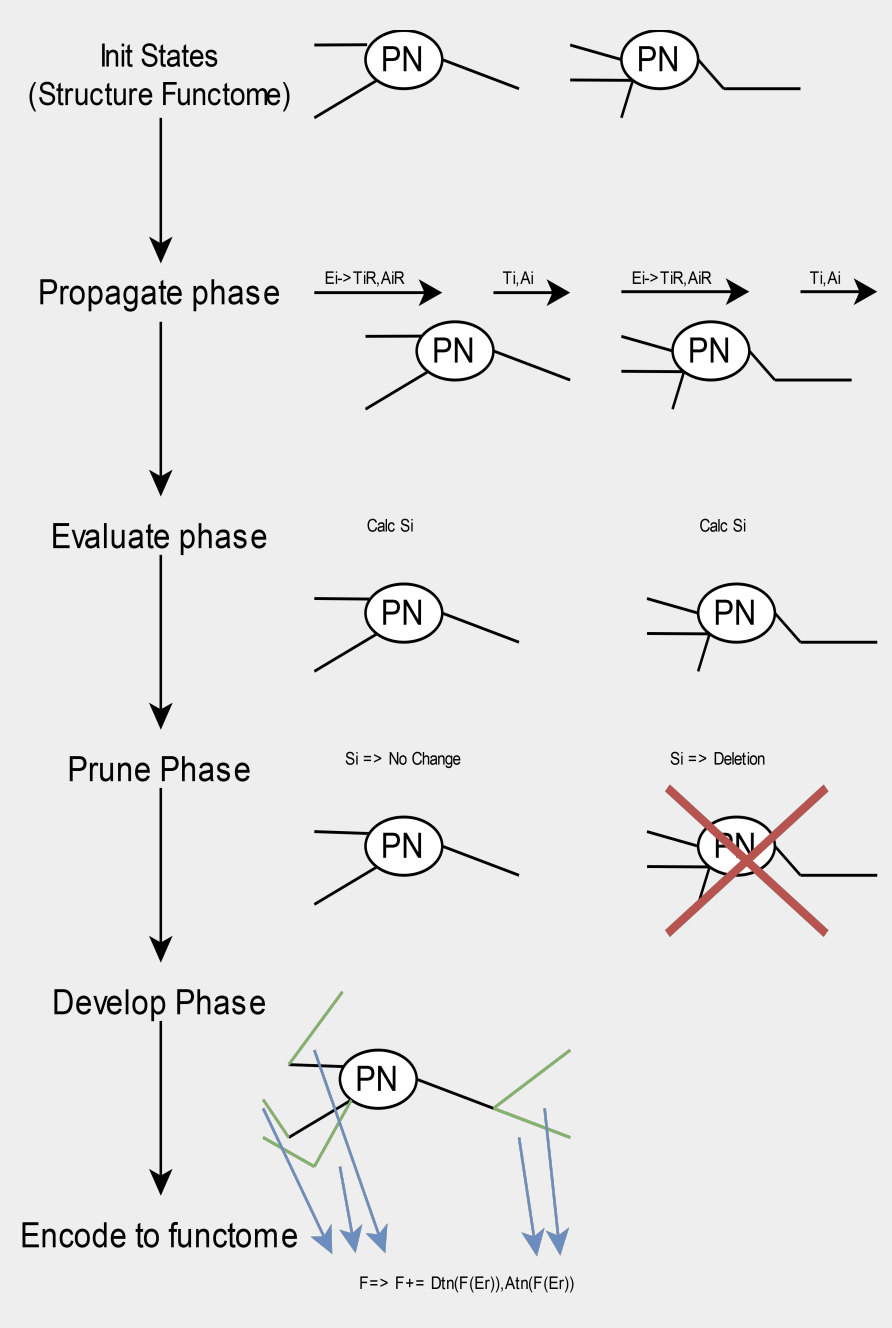

## GENETICS AND GENETIC ALGORITHMS

In a typical GA (Genetic algorithm) **[Ref 10]** We create a base definition (gene) that is partially or completely randomised. A generation of genes are then tested against the desired outcome and mutated. (A specialized Monte Carlo method) Generations are merged by a percentage and manipulated against the results measured; leading to hopefully an exponentially appropriate solution.

Whilst this method with enough Compute/Time will eventually minimize, seemingly minor flaws in the loss functions, selection criteria often cause significant waste and fragility of the solution and result in increasing risk.

In the biological variant, expression of the gene is also encoded in the genome, with the crucial difference of encoded behavior/dependencies of the expression. This creates functional hierarchies that lead to further expression and regulation. As a result, biological genetics do not suffer from over specialization to the point of brittle collapse under environment change due to linked dependencies and regulation in every encoded item; even with far more complex encoded behaviors over generations, only stable extensions to the base rule set that correctly operate previously successful regulation are maintained.

Encoded information is expressed based on feedback through the existing environment (external and internal/( In contrast GA's typically train within a narrow scope).

Post expression, manifested objects (E.G. Proteins) then operate within variance to also reinforce the environment expected of the genome, supporting further expression/regulation.

In contrast; standard GA's are severely limited compared to the biological which comprises of structure, growth and execution definitions, not simply randomized/mutated words.

For a more details on standard genetic algorithms please refer to **[Ref 14]**

## NATURE OF INFORMATION AND COMPLEXITY

### I. *Example in modern computing*

The binary standard 8 bit byte. From which more abstract types such as float or long integers are constructed.
Base types interact through a common rule set (Logical (bit-wise) or mathematical).
These processes are executed through registers which serve to perform ever higher abstractions through various languages.

All programming languages built on this architecture are interpretations and do not provide additional scope to the fundamental processes.

Notice that the bases of these types represent both base and structure, that is each bit of byte follows range for Br (Bit range) => 2^Bi (Bit index). This bit range is hierarchically dependent on the one preceding it.

### II. *How meaning is represented in compute*

The order of precedence/use of the hierarchical building blocks defines how meaning is translated to human context.

This precedent is contractually arbitrary and then optimized through hardware.

For example, a hard-drive typically stores less frequently accessed but more critically dependent information than RAM. (even more so with an L1 Cache).

Processing of meaning requires highest entropy components, and storing requires lowest entropy components – HDD/SSD (Where structured information is most dense).

### *III. Biological Neuron comparison*

*In contrast,* the brains most discrete transmission medium is an Ion. Whilst groups of ions can hold a variable charge unlike a binary hierarchy, their function within neurons is binary -operating ion gates ; this discrete action operating over analog thresholds of ion concentrations resemble a compute architecture:

III. **.Sodium and potassium ions** to regulate charge as a response from direct electrical or neurotransmitter excitation (triggering an in-balance).

**.Once a** charge differential between external and internal environment is beyond a given threshold, the neuron fires to the axon terminals, reinforced/regulated by the Myelin sheath produced by Shwann Cells; continuing the potassium/sodium propagation to final Calcium inflow and neurotransmitter[s] release or direct electrical stimulation.

**.The general i**ntensity of the stimuli is reflected in more frequent firings, where patterns of the stimuli reflect changes in the firing rate.

.**In memory** formation, groups of neurons grow to 'replay' memories of the past by generating the same collective output as before without the required chain of processed stimuli that created them -in other words an internal model, that is gradually more internally understood, (abstractedly similar to L1 to HDD process described in part 2 where clearly defined data structures are encoded).

**.The frequency** of firing reflects the neural coding of the stimuli. Various theories exist as to how this mapping encodes information precisely, however it is clear that the coding models the abstract information locally, with minor influences from global state; as apposed to back propagation in CNN systems: where error is translated to all layers. **Refs [8], [9] and [12].**

### IV. ***Entropy and criticality***

In both examples, points of high entropy correlate with least internally modeled information.

On the assumption that boundaries of hierarchical processes are defined by criticality of their operation. I propose the general rule applies in both nature and engineered computation: (Intuition section part IV:)

## Behavior driven minimization

In deep learning, data is trained with a defined objective within a coordinate space set.
The results of training is then interpreted externally, as either an interface for the system or to assist in refinement of the continued training.
The assumption in this method of training is that data of some domain contains useful features, these features share common traits and can be binned. These bins are defined as a product of the requested minimization; training creates micro translations of features and their representations into the output.

Whilst this method works for simple solutions through a network, training for global minimization leads to information loss that would otherwise be useful in assisting specific inferences.

All possible features of this complex domain are (assumed to be) within the model, however on testing a number of features happen to be contained within another, leading to less confidence or even the complete opposite output.
Whilst humans are also susceptible to this; a hierarchy of importance reduces this effect.

By training for network activity when some degree of domain related information is present, micro features can be trained and combined to provide a flexible and more reliable inference.

For example if someone where to ask 'Do you like my cat' you are then biased visually towards looking for one, narrowing the scope of search criteria in an entirely different domain before utilising the visual model, in other words; the application of a binned behavior is regulated by another.

Given this selection, the minimized output must be made aware to the more global 'selector' in order to determine the more appropriate response, this required relationship is similar to the architecture of a GAN **[Ref 16]**; whereby discriminator informs the generator how close it is to the real data input.
In contrast, an EDeN process node specialises to one or more stimuli by requiring less influence on the input dendrite terminals to produce the same spike for a given pattern (More on this later).

***Below depicts a high level control flow of this process:***

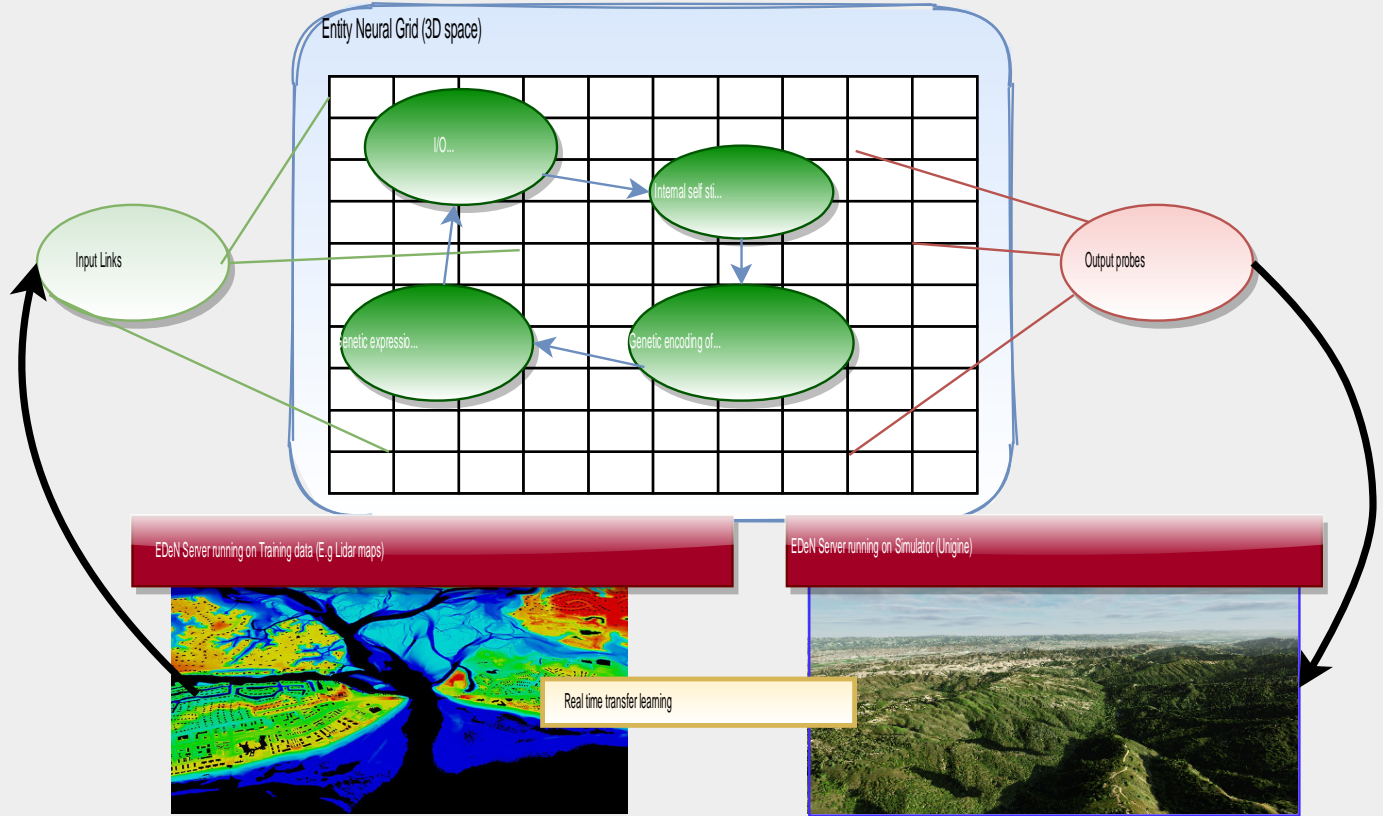

Viewer does not support full SVG 1.1

Cross correlation of supportive data from multiple domains at different frequencies, combined with local reinforcement of non random signals per process node (or neuron) then removes the requirement for specific global minimization goals.

In summary the pressure of the entities survival over time (more on stability index later) trains for specialisation per domain required of it.

Training the entity for a given task is to provide a reacquiring set of inputs interlaced with expected noise/variance.
The effect of an output probe on the input must then change the action to input to repeat itself within minimal variance in order for the entity to specialize.

Internally this forces both a model of the input/output relationship and a means to ignore irrelevant patterns.

### ARTIFICIAL AND BIOLOGICAL NEURONS

Artificial neurons work on the principle of a statically defined function/waveform that is then weighted at input set and singular output.
Results are evaluated through a loss function against the desired result from the end of the network; each neuron weight is adjusted to minimize the error against this output in relation to it's function [Layers examples: **Ref 11**] Neurons within hidden layers are minimized against the final output function relative to connected neurons. This encoding abstraction gradually trains a system to transform (and discriminate) data; applying back propagation through chains of partial derivatives.

Biological neurons attenuate their frequency of the coding pattern with no direct links to the required outcome [**Ref 7**]

This implies there is not a uniform function to each neuron (As with Specialized Deep learning Layers). But base rules of how morphology, and intra/extra cellular events regulate to produce this function intrinsically from local and global environment.

**.Activation functions**

**T**he common functions used in CNNs/Perceptrons operates in two dimensions and acts to exponentially decrease the effect of the weight summation beyond the mid-range values. In contrast to a biological neuron: both strength and frequency modulate **[Ref 12]** information. Neurotransmitter gradients, dendrite/axon interactions (dynamic growth and pruning) provide many more options for specialization.

**Global Propagation dependency**

The 'Hodgkin Huxley' model, changes weights based on the error of the models output and the desired output.

This first makes all relationships inside the network strongly coupled to the information structure of the output, error correcting based on the value local to the network and global. In other words translating/discriminating input into output. (CNN Layers equate to a complex convolution filter).

It is unclear to what degree biological neurons are directly dependent on their surroundings, however the myriad of studied morphology/genetic dependent processes suggest a more resilient model than global 'moment guidance' methods used today. **[Ref 13 ].**

**.Historical Neuron encoding**

A Hodgkin Huxley and common spiking neural models don't encode the history of activation, they are updated iteratively to the immediate weight model.

In order to retain classification across multiple Input outputs, network models must therefore generalise results.

**.Energy routing**

The EDeN Framework works to route 'energy' (a value that propagates over via Architectors and exchanged to the CEM , influencing energy propagation).

Multiple execution passes build energy values internal to a neuron (process node), while other process node functions regulate this behavior. This allows for multi variate processing based on both external and internal state, (a kind of currency exchange).

## Framework and core process overview

**The Neuron model**

Inspired by Self information theory. [**Ref 6**]. Data is received from the training environment and inflicts instability on a the neuron model, stimulating morphological response, leading to an incremental improvement in the minimization and internal modeling of data.

The neuron produces an output after building a minimal energy value via internal weights and routers. These act to minimize a stability index to acceptable threshold values.

The morphology of the neuron model is represented as vector locations of the dendrites and axon terminals, this produces delays and transformations in signal propagation by exchanging energy with 'Transmitters' or 'Architectors' that are analogous to neurotransmitters and neuropeptides.

**.The Neuron model: Update Method (Process node IO):**

On DevelopNetwork() function call, Process Nodes adjust their models using the following variables

**Ng** = Neural Grid

**Ei** [E, v(XYZ)]= Energy at an input location To the Ngrid (Neural Grid); delivered by an input probe.

**CEM** = Currently Energy Model (As a product of all neurites and soma process)

***The CEM utilises the following components:***

***T***[…]: An array of Axon Terminals

On activation, they release a Transmitter index payload as a Functome biased response from the EnergyValue.

**D**[…] An Array of Dentrites – these provide regulation to the CEM influence

**Tt/*At****: Transmitter/*Architector type, An Index and properties of the type if used by a Dentrite or Axon. (defined as TransArch payloads in code)

Variables unique to each specialisation determine the stimulation provided by the transmitter or morphology influence from architecture.

(This is designed to emulate the effect of ligand gates ION channel open/Closing on stimuli) and influence LTP like behaviour (long term potentiation).**Gc**[…]: An Array of Growth cones for each Dentrite or Axon that are regulated by the Functome expression.

**MiE** Minimum energy store before a spike, a consistent misfiring (or suppression) of the neuron due to forced response from extremes of energy will cause the stability of the neuron to decrease. (A tolerance of which causes pruning under phase control), this will result in random firings which do not filter out noise between correlated patterns observed by dendrite terminals.

Each Dentrite receives at least one TransmitterType, this type has a response that either blocks or allows energy updates to the process node.

'Architectors' are released based on measured acceptance criteria of a mutating Functome to provide structured improvement to process node morphology.

-Referenced as 'TransArch' Payloads in the framework.

**PWR 3D Tensors, (**Propagation, Weight and Router)

Work to produce spikes across each Z axis value, training for spike response from input sets per process node (See more information under CEM processing section)

**Generation/Discrimination states:**

**Generation** occurs when the Process Node model requires less input stimulus to produce a previously encoded response than that required to produce the original pattern. *(This process begins to occur post spike when the route and related weight matrix has been correctly enforced).*

This mechanism is also inspired by the prion theory of memory 'playback' by which only a fraction of the abstracted stimuli is required to reproduce the same signals.

**Discrimination** – the Process Node receives more inputs over an epoch than it was previously trained for, (stability index is unchanged or decreases), the Process Node is adjusting to new patterned stimuli and attempts to incorporate it into its existing model.

*High level diagram of forward propagation through the* ***CEM****:*

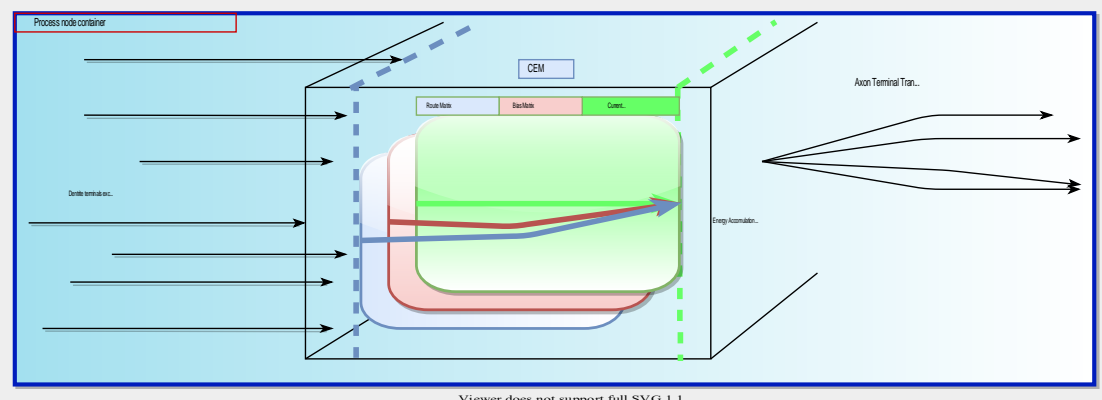

**Current Energy Model (CEM) processing**

The CEM acts to tune different stimuli to reinforce (or weaken) E at P[z].Given the propagation, weight and router Tensors (**PWR**) of dimensions X*Y*Z, where X == Y and Z is N, (N is user defined constraint), energy value E of a CEM where n is z index of an XY across the tensors.

Depicted right defines how PnE (Process node energy value) is determined.

A recursive algorithm across z of Z until PnE of the z plane is above the minimum threshold.

This triggers axon terminal firing which regulates this value by TransArch exchange to the external Neural grid, at which point all propagation values are reset to 0,

All Axon terminals are executed as immediately releasing transArchPayloads to the Neural Grid .

*Note: Other regulating mechanism may be applied in the future, as influence by homeostatic regulation in biology [**Ref 20**]*

Each router value (See R0 for router options) variable at n-1 xy plane on spike is updated to directly route to the spiking n plane if spike propagation energy (PE) is above Minimum Energy (MinE). Otherwise the value is set to a random variable of the Router options (E(R0)).

$$RO_x = \begin{matrix} \square & Ro_{010} & Ro_{020} \\ \square & Ro_{011} & Ro_{021} \\ \square & Ro_{012} & Ro_{022} \end{matrix}$$

$$RO_y = \begin{matrix} \square & Ro_{110} & Ro_{120} \\ \square & \square & Ro_{121} \\ \square & Ro_{112} & Ro_{122} \end{matrix}$$

$$RO_z = \begin{matrix} \square & Ro_{210} & Ro_{220} \\ \square & Ro_{211} & Ro_{221} \\ \square & Ro_{212} & Ro_{222} \end{matrix}$$

$$R_{xyz} \in RO$$

$$PE_n = \sum_{z=0}^{z=n} \{ At(PnE_{z-1}) + \sum_{x=0, y=0}^{x=X, y=Y} P_{xyz} \}$$

$$R_{xyz} \in \{0,1\}$$

$$W_{xyz} = [0,1]$$

$$P_{xyz} = W_{xyz} R_{xyz}$$

$$PNSD = \{ x > 0 | x \in \mathbb{R} \}$$

$$DR = \{ x > 0 | x \in \mathbb{R} \}$$

**The result of this process**

Rrouted weighted values are summed across the XY plane, minus decay, until the minimum spike energy is met. If this threshold is reached, the Process Node spikes and the successful local route is reinforced. If not, the route remains variable. In this way, the CEM determines whether a given internal energy pattern is sufficient to produce a stable spike response, and updates the local routing accordingly.

$$Assuming \sum_{i=0}^{|PNSD|} (PNSD-1_i - PNSD_i) \geq 0$$

$$PN_{SI} = \frac{1}{max((KL(PN_{SD-1} | PN_{SD})), 1)}$$

$$R_{xyz,n} = \begin{Bmatrix} E(RO), (PE_n < MinE) \\ R_{xyz,n-1}, (PE_n \geq MinE) \end{Bmatrix}$$

$$Process\ node\ internal\ routed\ transfer:$$

$$P_{xyz,n} = At(PE_{n-1}, R_{xyz,n}) W_{xyz,n}$$

$$PE_n = \sum_{x=0, y=0}^{x=X, y=Y} P_{xyz,n} - DR$$

$$S_n = \begin{Bmatrix} 1, (PE_n \geq MinE) \\ 0, (PE_n < MinE) \end{Bmatrix}$$

On forward propagation, routed weighted values are summed across the XY plane and the decay rate (DR) is then subtracted. This has the effect of having a 'cost' for propagation and only allowing sufficiently reinforced forward propagation to reach the spiking z. Unlike deep learning methods, the routed transfer receives only one active input per location, dictated by the current router state. Once a spike occurs during an epoch, the Process Node adjusts the local propagation and weight biases from the spiking Z position, such that routed dendrite influences that contributed to the spike may be reinforced positively or negatively.

Neurotransmitter and neuropeptide release. Given a complete picture of neuropeptide regulation in biological systems is not yet understood, the frameworks 'Architectors' provide a loose analogy, in that they regulate long term morphological impacts as opposed to the immediate effects of neurotransmitters. Current research suggests neuropeptides are managed by two key factors:.Production from active genes based on the developmental stage of the organism.Released when the state of the host neuron undergoes non standard spiking from their average.As such, if a spike occurs ,a neurotransmitter index is released, with a check of the spike frequency to also release an Architector. 'Actions' are then executed based on Architectors properties received by the post synaptic dendrite.Additionally, a hash exists per unique process node to help provide explicit characteristics and traceability of successful morphological configurations which are a result of allowed 'Action' activation (an analogy to active genes per cell type).Spike goal back propagation creationUnlike Deep learning; the goal is created dynamically on Spike in reference to the position of the spike along the Z axis, where the target is the propagation values at the last Z spiking plane. If the Process node measured stability drops below a threshold, the back prop goal is updated to the Z spiking plane propagation values of the current spike.Whilst the CEM tensors act as classical Deep Learning/Perceptron 'neurons', the router matrix configuration requires only one error calculation, as opposed to the common fully connected layer configuration requiring many, in this respect the process is more closely modelled to a biological equivalent, computationally. Updating the Stability index of the process node (PNSI) involves calculating the KL divergence across the previous Spike distribution (SD).

This has the effect of reinforcing the particular router configuration to spike at the same z position in the future, but requiring less input from the dendrite terminals in order to maintain the same z firing position (Otherwise stability index drops) Meanwhile any other inputs that are received in different phases are propagated through first the previously reinforced router/weight configuration up to z, then beyond (As they may not reach the minimum energy threshold).

**Input/Output probes**

To avoid a globally propagated minimization target that would enforce bias over the processing units localized development, Inputs act as a standard vector update fields. Whereas output probes are vector readers that are used in training or monitoring as part as physical entity simulation – with the intention of changing inputs in turn through, for example 'muscle' control (External to the Neural grid processes).

Inputs that have repetition then increase the stability indexes across all neurons; forcing adaption or 'death'.

**Genetic model and Functome overview**

*Authors note:*

*The right abstraction to take in genetic representation was first motivated by how a protein's 'process' is encoded (through Amino acids → RNA expression..) into DNA.*

*I decided against this due to the same behavior being plausibly expressed by*

*morphology of the neuron and variation in signal model and vector based adjustment rather than computationally expensive micro instructions.*

The Functome acts to encode behavior during all stages of entity development as options for future re-expression.
Each type and sub type within the framework contains a Functome reference.
Examples include positions for the initial structure of an entity and action options.
Generations of Functome expression lead to increased adaptability of the process node architecture under different input conditions.

This improves both the regulation of axiom expression and the structure of the network in relation to the overall ability of the network to handle unpredictable inputs.
An analogy to this methodology is to compare human and ape's language ability: Humans clearly have a genetic affinity to the general architecture required of speech, which is then specialised at approximately the same age.

***.Functome Encoded actions***
As the develop network phase is activated, 'AvailableActions' per Process node are scanned for by checking a prerequisite name and property constraints.
The mutation update modifies a collection of these commands per process node, references of this data is then stored in the Functome.

**Action types:**
.AxonTerminal_AddNew
.AxonTerminal_RemoveRandom
.Dentrite_AllowNew
.Dentrite_RemoveRandom
.AllowTransmitterIndexProduction
.AddArchitectorIndexProduction
.StimulateNeuroGenesis
.Apoptosis
(and more)
**Prerequisite type names:**

**.ArchitectorPresent**
**.TransmitterPresent**
**.ArchitectorPresentPayloadCount**
**.TransmitterPresentPayloadCount**
.AllowOn_ProcessNodeClockRange
.AllowOn_ClockFrequency
.EnergyRequirement
.EnabledAfterEntityClock
(and more)

Each Action is executed on the 'Develop' phase given perquisite type and it's meta data matches for a given action, or on Architectors received by a given process node at the propagation stage.

***. Initial growth structure and base growth***
We start with random available actions created per NeuralContainer, with random positions for cell body, axons and dendrite terminals encoded to the Functome.
On each epoch, neural containers execute these growth patterns, achieving changing stability index over time. Actions continue to be evaluated and mutated during the develop phase of the execution. This in combination with following existing growth rules slowly adapts the morphology of the process node.

***. Growth regulation strategy***
Much like natural neurons, synaptic targets must be regulated in order to build functional circuits; in that a single processNode ensures it is has grown it's dendrites into the a location where it can receive enough stimulation to succeed in generating a stable frequency.
To simulate this behavior, all dendrites have a limited number of growth iterations, with a 'grown cone' direction.
During development phase, the largest stimulating source is determined per dendrite, and the growth cone direction is adjusted to target that origin location when a spike occurs, on spiking; the log of origin stimulation is reset.

Axons on the other hand have a predisposed genetic target which is then influenced by 'Architectors' with a genetic bias.

This promotes the general Hebbian rule of 'wire together fire together'

## Architectors

Similar to neurotransmitter, 'Architectors' (An analogy to neuropeptides) exist to create new Process nodes and other static influences on the neural grid given specific prerequisites that apply to all other Neurites. *(Frequency and Density of Transmitter Indexes that provide discrete energy updates to the internal state of the unit)*. Both Initial Axioms and Functome definitions provide details as to what Process nodes are produced in terms of their activated actions. That is the assumption of the requirement to a given Neurite configuration before the environment argues for it's existence against normal Growth cone calculations. In contrast to Deep Learning or other static network definitions. This provides an element of 'disposability' to each Process node.

A requirement of this development exists in [**Ref 13**], whereby higher abstraction in the visual domain correlates less with direct stimuli, however it's unknown how this pattern projects to other existing biological domains or if what the boundary conditions of dimension returns are.

# Execution Phases Per Epoch

### *Propagate network*

Output probes and input links propagate energy updates via transmitter exchange to each entity (see section 'The Neuron model:' above for more details).

The CEM internally propagates it's energy updates and set's spike data if it occurs. This Spike data (Z index of the spike and total XY Plane energy) is used to tune the routers, update the current spike distribution, which is then used to back propagate the CEM weights for specialisation.

'TransArchPayloads' are then exchanged from CEM onto the neural grid based on the parent process nodes Functome encoded instructions.

## *Evaluate and Prune Network*

The Stability index is calculated for all process nodes, those that fall below the minimal threshold are removed – emulating neural pruning.

If the CEM of a process node hasn't been stimulated by any dendrite terminals, stability index and prune functionality is disabled for the epoch.

## *Develop Network*

The remaining Process nodes are allowed to be modified based on Evaluate Network results, new process nodes and Neurite updates are executed. The active Functome functions continue to propagate within the process nodes.

Mutations are applied randomly to the growth rules/actions if the stability index is below a set minimum.

# Temporal eligibility, delayed reinforcement and fitness attenuation

Immediate phase results alone do not explain whether a recent local route caused a later successful or unsuccessful outcome. As shown in [**Ref 21**], recent local activity may be retained as a short lived eligibility trace until later reward, punishment or other modulatory feedback argues for or against the same route.

As an extension to the Develop phase, recent Dendrite stimulation, Process Node activation and Axon Terminal release are retained as short lived eligibility traces, separate from current morphology and the Functome. Eligibility decay determines how long this recent activity is retained. Reward sensitivity, punishment sensitivity and structural credit bias determine how later feedback changes the same route.

Output Probe evaluation and other environmental feedback then produce signed temporal credit against the recent local process. This credit is consolidated as a temporary causal map over Dendrite Terminals, Process Nodes and Axon Terminals. Develop support, Prune pressure and Prune protection are then derived per local process.

In training, recent successful local routes receive increased pressure for retention or further expression, while weak, unstable or repeatedly unrewarded routes receive increased pressure for attenuation or removal. In relation to the Functome, this does not replace encoded behavior, but adds pressure as to which stable local behaviours are retained, weakened or further expressed.

Domain response time limits and fitness attenuation reduce repeated explicit enforcement once a domain is sufficiently encoded. This provides a means to associate delayed success or failure with recent local process, without requiring a global back propagation method. [**Ref 21**]

***All variables used in this regulation process:***

$DT = DendriteTerminal$
$PN = ProcessNode$
$AT = AxonTerminal$
$L \in \{DT, PN, AT\}$
$A = recent\ local\ activity$
$E = eligibility\ trace$
$OPpos = positive\ Output\ Probe\ result$
$OPneg = negative\ Output\ Probe\ result$
$F = delayed\ signed\ feedback$
$C = signed\ temporal\ credit$
$TC = consolidated\ temporal\ credit$
$D = Develop\ support$
$PP = Prune\ pressure$
$PT = Prune\ protection$
$EDec = eligibility\ decay$
$RDec = reward\ trace\ decay$
$TCDec = temporal\ credit\ consolidation\ decay$
$Rew = reward\ sensitivity$
$Pun = punishment\ sensitivity$
$DB = Develop\ bias$
$PB = Prune\ bias$
$SCB = structural\ credit\ bias$
$Low = low\ credit\ pressure$
$Instab = instability\ pressure$
$Under = chronic\ underuse\ pressure$

***Used as shown***

$$A_{DT,n} = Stm_{DT,n}$$
$$A_{PN,n} = S_n$$
$$A_{AT,n} = Rel_{AT,n}$$

$$E_{L,n} = EDec_L E_{L,n-1} + A_{L,n}$$
$$F_n = OPpos_n - OPneg_n$$
$$C_{L,n} = RDec_L C_{L,n-1} + \begin{cases} Rew_L F_n E_{L,n}, (F_n \geq 0) \\ Pun_L F_n E_{L,n}, (F_n < 0) \end{cases}$$
$$TC_{L,n} = TCDec_L TC_{L,n-1} + C_{L,n}$$

$$D_{L,n} = max(TC_{L,n}, 0) DB_L$$
$$PP_{L,n} = \{Low_{L,n} + Instab_{L,n} + Under_{L,n} + max(-TC_{L,n}, 0)\} PB_L$$
$$PT_{L,n} = max(TC_{L,n}, 0) SCB_L$$

Where L defines the local process under evaluation, where L ∈{DT, PN, AT}, corresponding to Dendrite Terminal, Process Node or Axon Terminal.

- A_{DT,n} Dendrite Terminal activity at step `n`.
- A_{PN,n} Process Node activity at step `n`.
- A_{AT,n} Axon Terminal release at step `n`.
- E_{L,n} Short lived eligibility of local process `L`.
- F_{n} Delayed signed feedback from Output Probe results.
- C_{L,n} Signed temporal credit of local process `L`.
- TC_{L,n} Consolidated temporal credit of local process `L`.
- D_{L,n} Develop support applied to local process `L`.
- PP_{L,n} Prune pressure applied to local process `L`.
- PT_{L,n} Prune protection applied to local process `L`.

At **Develop Phase**, `D_{L,n}` increases pressure for further local expression where recent local activity is later supported by delayed feedback. `PT_{L,n}` also helps prevent useful recent local structure from being lost too early.
At **Prune Phase**, `PP_{L,n}` increases pressure against weak, unstable or repeatedly poor local structure. `PT_{L,n}` acts in opposition by protecting recently useful local structure from premature pruning.

**In summary:**

- `D_{L,n}` supports further local development.
- `PP_{L,n}` supports attenuation or removal.
- `PT_{L,n}` protects useful recent local structure.

A copy of this save state can either be 'locked' - whereby only the propagate phase is activated or used for further development under potentially different environments or within a simulation/embedded context with other entities with alternative specialised.
This system encourages transfer learning as a core principle of development.

**Workflow description**

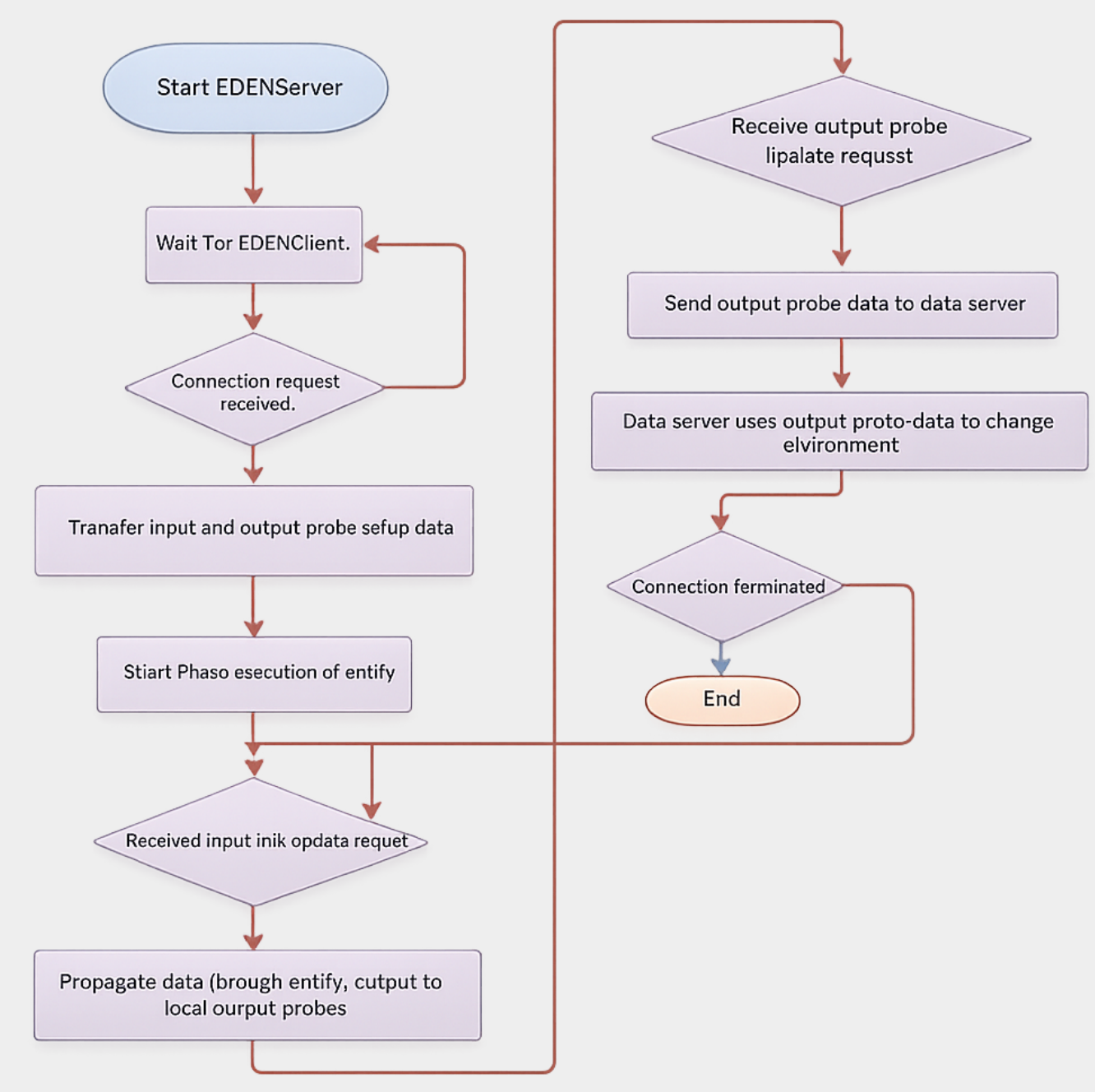

**Depicted above, a high level diagram of how data is propagated between variations of the EdeN Server and EdeN Client.**

## Notes on Entity Fitness Evaluation method

In order to evaluate the progression of entities over many domains simultaneously, a fitness function from both the explicit data of a given input function and it's potential cross domain outputs have to be supplied.
Naturally, the number of evaluations required approach the infinite after a very short duration.
There are a few strategies to deal with this issue:

I. Domain response time constraints, whereby a certain minimum and maximum cardinality of the events of an agent is required for it's survival.

II. Self attenuating fitness evaluation, whereby on a certain milestone of the multi-agent training, a domains performance is deemed 'good enough' for a reduction of the same evaluation to only be required, this assumes this same training data is so well encoded into the surviving agents, further explicit enforcement isn't required.

## Training evaluation overview

Alongside the creation of entities and it's evolving constraints, the EdeN Framework also provides a training guidance functionality, labeled as 'TrainingSessionMetaData' this is a runtime log of all entity virtual positions, value goals and virtual environmental control data which create or destroy elements in the training environment on acceptance of various preconditions. This allows for a dynamic environment aimed at promoting the evolution of highly complex tasks.

## High level project component diagram

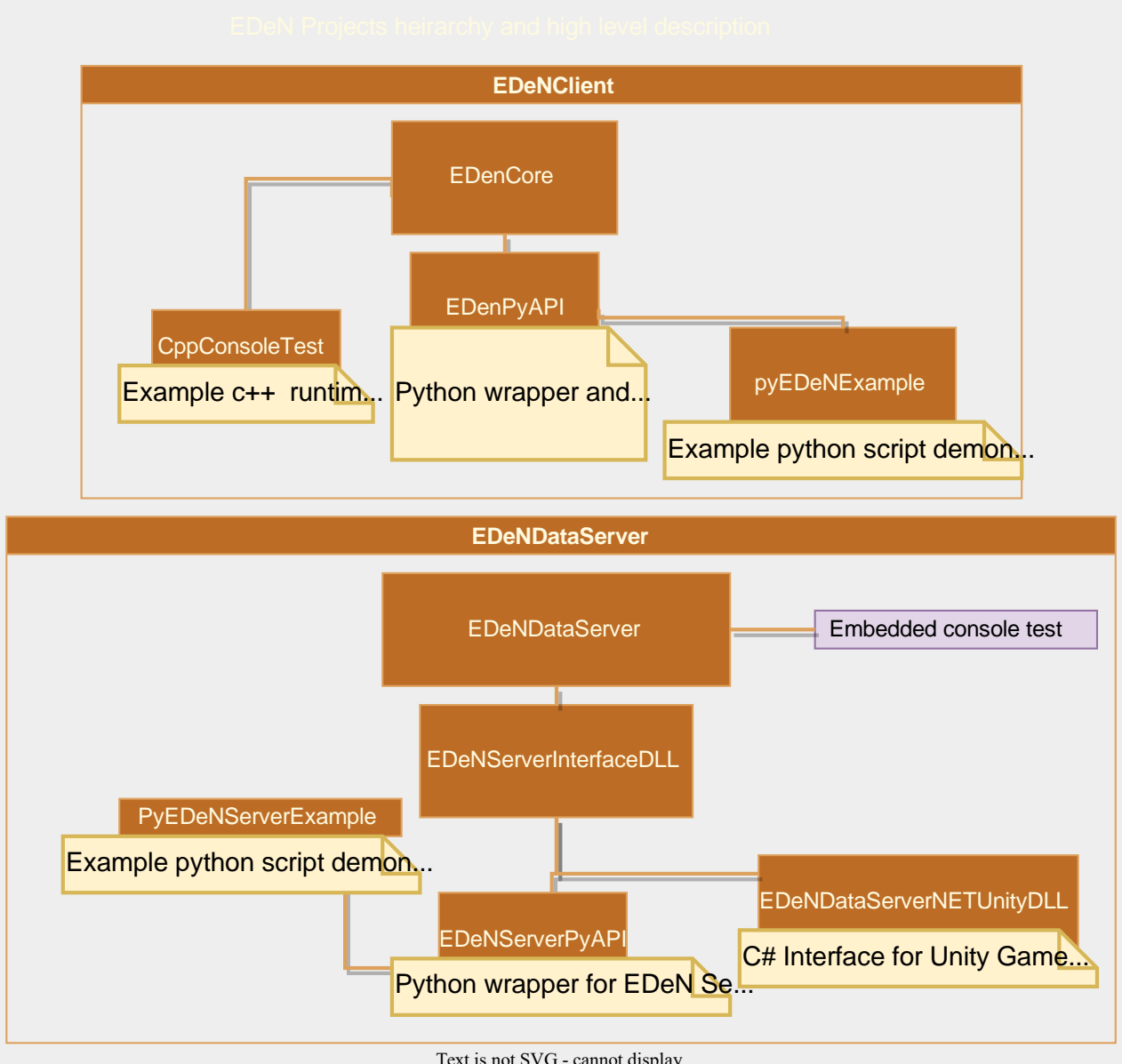

## Closing comments and next steps

Whilst all mechanisms described above function as expected within the framework, large scale testing for trained entities performing complex tasks has yet to be achieved, further research and development work continues in this area and more papers are to follow with results.

It is believed the methods selected in this framework will provide a better interface to general training without requiring specialist knowledge, leading to the authors desired goal of supportive/ duplicate systems to aide sectors from healthcare to R&D and manufacturing.

## References and Influential sources

### Authors

Jamie Nicholas Shelley, Optishell Consultancy